\documentclass[runningheads]{llncs}
\usepackage[T1]{fontenc}
\usepackage{graphicx}
\usepackage{makeidx}  
\usepackage{silence}
\let\vec\relax
\usepackage{amsmath}
\usepackage{amssymb}
\usepackage[colorlinks=true, allcolors=blue]{hyperref}
\usepackage[numbers]{natbib}
\usepackage{booktabs}
\usepackage{pdflscape}
\usepackage{adjustbox}
\usepackage{array}
\usepackage{svg} 
\usepackage{multirow}
\usepackage{makecell}
\usepackage{multicol}
\usepackage{tabularx}
\usepackage{arydshln}
\usepackage{bm}
\usepackage[caption=false]{subfig}
\usepackage{enumitem}
\newcommand{\mypar}[1]{\smallskip\noindent\textbf{#1.}}
\newcommand{\mypartwo}[1]{\vspace{0.5pt}\noindent\textit{#1.}}

\begin{document}
\title{Revisiting Predictive Process Monitoring in the Age of Foundation Models: A Comparative Study of Sequence, Tabular, and LLM Approaches}
\titlerunning{Revisiting PPM Foundation Models}
%

\author{Lennart Fertig\inst{1} \and
Lukas Kirchdorfer\inst{1,2} \and
Tobias Sesterhenn\inst{3}
}

\authorrunning{L. Fertig et al.}

%
\institute{University of Mannheim, Mannheim, Germany\\
\email{lennart.fertig@students.uni-mannheim.de}
\and 
SAP Signavio, Walldorf, Germany\\
\email{lukas.kirchdorfer@sap.com}
\and
Technical University of Clausthal, Clausthal, Germany\\
\email{tobias.sesterhenn@tu-clausthal.de}
}
\maketitle              
\begin{abstract}
Predictive process monitoring (PPM) leverages event logs to forecast the future of running process instances, for instance, predicting the next activity, the remaining time until case completion, or the time to the next event. While PPM research in recent years has been dominated by deep sequence models trained from scratch, such as Long Short-Term Memory (LSTM) models, foundation-model approaches---particularly large language models (LLMs)---are increasingly explored for PPM. At the same time, tabular foundation models with in-context learning capabilities offer a promising alternative but have not yet been systematically benchmarked for PPM.  Thus, it remains unclear whether classical sequence-based models remain competitive in this evolving landscape.
This paper compares the three modeling paradigms both conceptually and empirically through a controlled benchmark across multiple datasets and prediction tasks. 
The results show that sequence models consistently perform best for next activity prediction, whereas tabular foundation models are competitive on temporal tasks, with LLMs usually lagging behind despite higher cost.


\keywords{Predictive Process Monitoring  \and Large Language Models \and Tabular Foundation Models.}
\end{abstract}
\section{Introduction}
\label{sec:intro}
Predictive process monitoring (PPM) is a subfield of process mining that leverages historical process execution data---typically stored in event logs---to predict future properties of individual, currently running process instances. Typical tasks include predicting the next activity (NA), the remaining time until completion of the case (RT), and the time until the next event (NT)~\cite{bertiInContextFoundationModel2026a,wangMTLFormerMultiTaskLearning2023}. 
These capabilities are relevant in complex operational settings, e.g., in supply chains, multi-stage approval workflows, or service processes with strict time constraints, where early warnings about delays or bottlenecks can trigger corrective actions~\cite{tanikondaApplicationTransformerModels2022}.

In recent years, PPM research has been dominated by deep sequence models that are trained from scratch for a specific process, most prominently in the form of recurrent neural networks (RNN) such as Long Short-Term Memory (LSTM) and Transformer-based models~\cite{rama-maneiroDeepLearningPredictive2023,wangMTLFormerMultiTaskLearning2023}. More recently, pre-trained foundation models have been explored for PPM, including large language models (LLMs) adapted to structured event log representations~\cite{oyamadaDomainAdaptationLLMs2025}. In parallel, foundation models for tabular prediction with in-context learning have emerged beyond text~\cite{hollmannAccuratePredictionsSmall2025,spinaciConTextTabSemanticsAwareTabular2025a}, motivating the question of whether such models can serve as competitive baselines when event log prefixes are represented in tabular form.

However, only limited attention has been given to systematically comparing foundation models with classical sequence models for PPM tasks. Oyamada et al.~\cite{oyamadaDomainAdaptationLLMs2025} take an important step in this direction by comparing fine-tuned LLMs with sequence-based baselines for NA and RT prediction. Nevertheless, several questions remain open for the process mining community. First, a conceptual comparison of the underlying modeling paradigms is still missing. Second, tabular foundation models have, to the best of our knowledge, not yet been evaluated for PPM, despite their promising performance in other domains. Third, there is limited understanding of when and why different paradigms succeed or fail across PPM tasks and event logs.

To address this gap, this paper makes three main contributions. First, we conceptually compare three modeling paradigms---classical sequence models trained from scratch, LLM-based approaches adapted via Low-Rank Adaptation (LoRA), and tabular foundation models using in-context learning---along the dimensions of prefix representation, backbone parameterization, and adaptation regime. Second, we conduct a controlled empirical benchmark across five real-world event logs and three prediction tasks (NA, RT, NT), comparing predictive performance and runtime. Third, we provide insights into performance differences, focusing on branching-related prediction difficulty and selected model-specific failure patterns.

\section{Background and Related Work}
\label{sec:rel_work}

\subsection{Predictive Process Monitoring}
\label{sec:rw_ppm}

PPM leverages historical event logs to forecast the future evolution of running process instances and to enable proactive decision-making through timely interventions~\cite{difrancescomarinoPredictiveProcessMonitoring2022}.
Common prediction targets include NA, RT, and NT~\cite{vanderaalstProcessMiningHandbook2022,wangMTLFormerMultiTaskLearning2023}. 
Earlier work in PPM includes state-based and classical machine learning approaches that often depend on explicit feature engineering, such as transition-system-based methods and other shallow predictors~\cite{vanderaalstTimePredictionBased2011,polatoTimeActivitySequence2018,appiceLeveragingShallowMachine2019}. Subsequent research increasingly adopted deep learning architectures to reduce reliance on manual feature construction and to model sequential dependencies in event logs~\cite{lecunDeepLearning2015,kratschMachineLearningBusiness2021}.
Building on this, RNNs, particularly LSTM variants, have been applied to next activity prediction, next event timing, and remaining time estimation~\cite{taxPredictiveBusinessProcess2017,camargoLearningAccurateLSTM2019}. Complementing RNN and LSTM variants, convolutional neural networks (CNNs), including stacked-inception CNN modules, have been studied for next-activity prediction and compared against LSTM-based models~\cite{dimauroActivityPredictionBusiness2019, wangMTLFormerMultiTaskLearning2023}. Di Mauro et al.~\cite{dimauroActivityPredictionBusiness2019} report higher accuracy than an LSTM baseline across all evaluated datasets. More recently, attention mechanisms have been integrated into predictive models, while Transformer-based architectures have been introduced to capture long-range dependencies in event sequences~\cite{wickramanayakeBuildingInterpretableModels2022,bukhshProcessTransformerPredictiveBusiness2021}. Bukhsh et al.~\cite{bukhshProcessTransformerPredictiveBusiness2021} report that ProcessTransformer outperforms several prior baselines for next activity prediction and remains competitive for next event timing and remaining time estimation.

\subsection{Foundation Models in Process Mining}
\label{sec:foundation_models_in_PM}

Foundation models are pretrained on broad datasets and can be adapted to downstream tasks with limited task-specific training. 
Motivated by this shift, recent work argues that process mining, including PPM, can benefit from foundation-model capabilities, such as reusable representations and efficient adaptation to new tasks and datasets~\cite{bertiInContextFoundationModel2026a}.

Within process mining, LLMs have been explored for a broad range of tasks due to their strong general-purpose pattern learning and instruction-following capabilities~\cite{bertiPMLLMBenchmarkEvaluatingLarge2024,brownLanguageModelsAre2020}.
For PPM in particular, two main approaches have emerged: \emph{textual abstraction}, which transforms event logs into natural-language descriptions and queries LLMs via prompting~\cite{kubrakExplanatoryCapabilitiesLarge2024,bertiPMLLMBenchmarkEvaluatingLarge2024,pasquadibisceglieLUPINLLMApproach2024,rebmannEvaluatingAbilityLLMs2024a}, and \emph{structured adaptation}, which fine-tunes LLMs on event-log representations using parameter-efficient methods such as LoRA~\cite{huLoRALowRankAdaptation2021,oyamadaDomainAdaptationLLMs2025}.
Our empirical comparison follows the structured approach, as it has shown stronger performance than prompting-based methods~\cite{oyamadaDomainAdaptationLLMs2025}.

In parallel to LLMs, tabular foundation models with in-context learning capabilities have emerged for tabular prediction tasks~\cite{hollmannAccuratePredictionsSmall2025,spinaciConTextTabSemanticsAwareTabular2025a}.
Unlike sequence models, these approaches operate on fixed-length feature vectors and require no gradient-based training at inference time.
However, to the best of our knowledge, they have not yet been systematically evaluated for PPM tasks.

The most closely related work is by Oyamada et al.~\cite{oyamadaDomainAdaptationLLMs2025}, who evaluate fine-tuned decoder-only LLMs for PPM and compare against established baselines, including LSTM and Transformer models, under leakage-avoiding temporal splits.
Their results indicate that fine-tuned LLMs can be competitive with classical sequence models on NA and RT prediction.
However, as argued in Section~\ref{sec:intro} this benchmark remains incomplete for a comprehensive comparison of modeling paradigms in PPM, which motivates our unified benchmark comparing classical sequence models, LLM-based approaches, and tabular foundation models across three PPM tasks.
\section{Preliminaries}
\label{sec:prelim}

\mypar{Event log, traces, prefixes}
Process execution data is assumed to be stored in an event log $\mathcal{E}$, defined as a finite set of traces. Each trace $\sigma = \langle e_1, e_2, \ldots, e_n\rangle$ represents the ordered sequence of events for a single process instance (case). An event is defined as a tuple $e=(c,a,r,t)$, consisting of a case ID $c$, an activity $a$, a resource attribute $r$, and a timestamp $t$. We use the projection functions $\pi_A(e)$ and $\pi_T(e)$ to retrieve the activity and timestamp of an event, respectively.

PPM is typically formulated as a classification or regression problem and begins with a feature extraction step. This involves generating prefixes of varying lengths from each completed trace to represent different execution stages. Formally, for a given trace $\sigma$, its prefix of length $k \in {1,\ldots,n-1}$ is defined as $p_k(\sigma)=\langle e_1,\ldots,e_k\rangle$, representing a partial execution up to the $k$-th event of a trace with $n$ events. Each prefix is mapped to an input representation $\bm{x}\in\mathcal{X}$ and associated with a task-specific target value $y_\tau\in\mathcal{Y}_\tau$, where $\tau\in\{\mathrm{NA},\mathrm{NT},\mathrm{RT}\}$ denotes the prediction task.

\mypar{Definition of PPM tasks}
We consider three PPM tasks. Each task maps a prefix $p_k(\sigma)$ to a task-specific target value:

\begin{enumerate}[noitemsep,topsep=0pt]
    \item \textbf{Next activity prediction}: Defined as
    $\Theta_{\mathrm{NA}}(p_k(\sigma))=\pi_A(e_{k+1})$,
    predicting the activity of the next event.

    \item \textbf{Next event time}: Defined as
    $\Theta_{\mathrm{NT}}(p_k(\sigma))=\pi_T(e_{k+1})-\pi_T(e_k)$,
    predicting the time difference between the next event and the current event.

    \item \textbf{Remaining time prediction}: Defined as
    $\Theta_{\mathrm{RT}}(p_k(\sigma))=\pi_T(e_n)-\pi_T(e_k)$,
    predicting the time difference between the final event of the case and the current event.
\end{enumerate}
\section{Modeling Paradigms for Predictive Process Monitoring}
\label{sec:models}
This section compares three modeling paradigms for PPM: 
(1) classical sequence models, (2) LLM-based approaches, and (3) tabular foundation models. \autoref{tab:comparison} provides a high-level overview.

Let $p_k(\sigma)=\langle e_1,\dots,e_k\rangle$ denote a prefix and 
$\Theta_\tau(p_k(\sigma))$ the target for task $\tau\in\{\mathrm{NA},\mathrm{NT},\mathrm{RT}\}$ as defined in Section~\ref{sec:prelim}.  
All paradigms instantiate a predictor
$f_\tau : \mathcal{P} \rightarrow \mathcal{Y}_\tau$
mapping prefixes to task-specific outputs.
They differ along three main dimensions:  
(i) prefix representation,  
(ii) backbone parameterization, and  
(iii) learning or adaptation regime.  

Classical sequence models and LLM-based approaches share the same sequential prefix representation and differ primarily in backbone initialization and training regime.  
Tabular foundation models instead represent prefixes as fixed-length feature vectors and treat PPM as tabular prediction.

\subsection{Classical Sequence Models}

\mypar{Prefix representation}
Classical sequence models encode a prefix as an ordered sequence of event embeddings.  
Let
$\phi_{\mathrm{seq}} : \mathcal{E} \rightarrow \mathbb{R}^d$
map an event to a $d$-dimensional embedding by combining categorical embeddings (e.g., activity, resource) and projected numerical attributes (e.g., timestamps).  
A prefix is represented as
\[
\mathbf{h}_{1:k}
= \langle \phi_{\mathrm{seq}}(e_1),\dots,\phi_{\mathrm{seq}}(e_k)\rangle
\in (\mathbb{R}^d)^k,
\]
preserving execution order and temporal dependencies.

\mypar{Backbone and predictor}
The predictor is parameterized as
\[
f_\tau(p_k(\sigma);\theta,\omega_\tau)
= g_{\tau,\omega_\tau}\!\left(\mathrm{BB}_{\theta}(\mathbf{h}_{1:k})\right),
\]
where $\mathrm{BB}_{\theta}$ is a sequential backbone (e.g., LSTM or Transformer), 
$g_{\tau,\omega_\tau}$ is a task-specific head, and 
$\theta$ denotes the backbone parameters, and $\omega_\tau$ denotes the parameters of the task-specific head.

\mypar{Adaptation regime}
All parameters are learned from event log data via supervised optimization:
\[
\begin{aligned}
\theta^* = \arg\min_\theta
\sum_{p_k(\sigma)\in\mathcal{D}}
\mathcal{L}_t\!\bigl(
f_t(p_k(\sigma);\theta),\,\Theta_t(p_k(\sigma))
\bigr).
\end{aligned}
\]
Both the event embedding $\phi_{\mathrm{seq}}$ and the sequential dynamics are inferred solely from process data.

\begin{table}[t]
\centering
\caption{Conceptual comparison of PPM modeling paradigms.}
\begin{tabularx}{\linewidth}{l>{\raggedright\arraybackslash\hsize=0.95\hsize}X>{\raggedright\arraybackslash\hsize=1.25\hsize}X>{\raggedright\arraybackslash\hsize=0.8\hsize}X}
\toprule
Dimension & Sequence models & LLM-based & Tabular FM \\
\midrule
Prefix representation & event sequence & event sequence & feature vector \\
Backbone & $\mathrm{BB}_{\theta}$ & $\mathrm{BB}_{\theta_0,\psi}$ & TFM \\
Pretraining & none & sequence pretraining & tabular pretraining \\
Training on log & full $\theta$ & $\psi$ only & none \\
Adaptation mechanism & supervised & PEFT & in-context \\
\bottomrule
\end{tabularx}
\label{tab:comparison}
\end{table}

\subsection{Large Language Model--Based Approaches}

\mypar{Prefix representation}
LLM-based approaches preserve the same ordered prefix structure as classical sequence models, but employ a different event encoding strategy that aligns process events with the input space of a pretrained Transformer backbone.

\mypar{Backbone and predictor}
The predictor reuses a pretrained backbone $\mathrm{BB}_{\theta_0}$:
\[
\begin{aligned}
f_t(p_k(\sigma);\theta_0,\psi)
&= g_{t,\psi}\!\left(\mathrm{BB}_{\theta_0,\psi}(\mathbf{h}_{1:k})\right),
\end{aligned}
\]
where $\theta_0$ are frozen pretrained weights and $\psi$ are adaptation parameters (e.g., LoRA adapters or task heads).

\mypar{Adaptation regime}
LLM-based approaches can be applied through prompting-based zero-shot or few-shot inference, or adapted to process data through fine-tuning. In this work, we use parameter-efficient fine-tuning (PEFT) via LoRA, where only $\psi$ is optimized while $\theta_0$ remains fixed:
\[
\begin{aligned}
\psi^* = \arg\min_\psi
\sum_{p_k(\sigma)\in\mathcal{D}}
\mathcal{L}_\tau\!\bigl(
f_\tau(p_k(\sigma);\theta_0,\psi),\Theta_\tau(p_k(\sigma))
\bigr).
\end{aligned}
\]
This adapts the pretrained backbone to process data without retraining its full parameter set.

\subsection{Tabular Foundation Models}

\mypar{Prefix representation}
Tabular foundation models map each prefix to a fixed-length feature vector:
\[
\phi_{\mathrm{tab}} : \mathcal{P} \rightarrow \mathbb{R}^m,
\qquad
\mathbf{x} = \phi_{\mathrm{tab}}(p_k(\sigma)).
\]
Features summarize prefix properties such as current activity, elapsed time, or positional indicators.  
Since the model operates on rows rather than sequences, ordering and temporal dependencies must be encoded explicitly. Tabular foundation models therefore replace the sequential representation with a feature-based state representation, modeling process dynamics through engineered features rather than an explicit event sequence.

\mypar{Backbone and predictor}
For each task $\tau$, let
\[
\mathcal{T}_\tau
=
\left\{
\left(
\mathbf{x}_i,
\Theta_\tau\!\left(p_{k_i}(\sigma_i)\right)
\right)
\right\}_{i=1}^{N}
\]
denote the task-specific context set. A tabular foundation model with fixed pretrained parameters $\eta_0$ predicts via in-context learning:
\[
\begin{aligned}
f_\tau\!\left(p_k(\sigma)\mid\mathcal{T}_\tau;\eta_0\right)
&=
\mathrm{TFM}_{\eta_0}\!\left(\mathbf{x}_q;\mathcal{T}_\tau\right), \\
\mathbf{x}_q
&=
\phi_{\mathrm{tab}}\!\left(p_k(\sigma)\right).
\end{aligned}
\]

\mypar{Adaptation regime}
No gradient-based parameter updates occur at inference time.  
Adaptation arises solely through contextual conditioning on observed rows and their labels.
\section{Evaluation}
\label{sec:eval}
This section presents our benchmark of the different modeling paradigms for PPM. In the remainder,
Section~\ref{sec:setup} describes the experimental setup, before
Section~\ref{sec:results} reports the results. The benchmark implementation supporting reproducibility is available online\footnote{\url{https://github.com/Privajet/revisiting-ppm-foundation-models}}.

\subsection{Experimental Setup}
\label{sec:setup}
For our experimental setup, we largely follow Oyamada et al.~\cite{oyamadaDomainAdaptationLLMs2025}. Specifically, we use the same datasets, preprocessing, and data split.

\mypar{Datasets}
Five publicly available event logs are used from the 4TU.ResearchData repository: BPI12\footnote{\url{https://data.4tu.nl/articles/_/12689204/1}}, BPI17\footnote{\url{https://data.4tu.nl/articles/_/12696884/1}}, and three variants of the BPI20 collection (BPI20RfP\footnote{\url{https://data.4tu.nl/articles/_/12706886/1}}; BPI20PTC\footnote{\url{https://data.4tu.nl/articles/_/12696722/1}}; BPI20TPD\footnote{\url{https://data.4tu.nl/articles/_/12718178/1}}).
These logs cover substantial variation in log size, activity alphabet, and trace-length characteristics (see Table~\ref{tab:oyamada_log_properties}).

\begin{table}[t]
\centering
\caption{Event log properties.}
\label{tab:oyamada_log_properties}
\small
\setlength{\tabcolsep}{6pt}
\renewcommand{\arraystretch}{1.15}
\begin{tabular}{lrrrr}
\toprule
\textbf{Log} & \textbf{\# cases} & \textbf{\# evt.} & \textbf{\# act.} & \textbf{Trace length} \\
\midrule
BPI20PTC & 2099  & 18246   & 29 & $8.6927\pm2.3$ \\
BPI20RfP & 6886  & 36796   & 19 & $5.3436\pm1.5$ \\
BPI20TPD & 7065  & 86581   & 51 & $12.2549\pm5.6$ \\
BPI12    & 13087 & 262200  & 24 & $20.0351\pm19.9$ \\
BPI17    & 31509 & 1202267 & 26 & $38.1563\pm16.7$ \\
\bottomrule
\end{tabular}
\end{table}

\mypar{Data Split} 
We apply the leakage-avoiding temporal split proposed by Weytjens and De Weerdt~\cite{weytjensCreatingUnbiasedPublic2022}. Using a nominal 80/20 case-level split, cases are ordered by their first-event timestamp and any case overlapping the split point is assigned to the test set. Consequently, the realized ratio may slightly deviate from 80/20. During preprocessing, cases with fewer than two events are removed and numerical features are z-score normalized. Event sequences are encoded using trace encoding~\cite{roiderEfficientTrainingRecurrent2024}.

\mypar{Features}
Across all paradigms, we use the activity label as a categorical input feature. The numerical input features comprise accumulated case time, day of month, day of week, day of year, hour of day, minute of hour, month of year, second of minute, seconds within the day, and week of year.

\mypar{Benchmark Approaches}
We compare the three paradigms discussed in Section~\ref{sec:models}: \textit{sequence models} are trained from scratch, \textit{tabular FMs} use in-context learning without training, and \textit{LLMs} use LoRA fine-tuning.
All paradigms are evaluated on the same three prediction tasks: NA, RT, and NT. Sequence models and LLM-based approaches jointly predict the three targets in a multi-task setting, whereas the tabular FMs use separate task-specific predictors.

\mypartwo{Classical sequence baselines}
Two sequence models are trained from scratch per event log.
The recurrent baseline uses a tunable LSTM backbone, following an established LSTM-based PPM approach~\cite{taxPredictiveBusinessProcess2017}. The Transformer baseline uses a Transformer sequence model with learned positional embeddings and causal self-attention.

\mypartwo{Decoder-only LLM backbones}
Following Oyamada et al.~\cite{oyamadaDomainAdaptationLLMs2025}, we adapt LLM-based approaches to process data using a structured event-log representation. Activity labels are treated as process-native tokens rather than natural-language text.
We evaluate two decoder-only LLMs adapted via LoRA: Llama-3.2-1B and Gemma-2-2b.

\mypartwo{Tabular foundation model baselines}
We benchmark two tabular foundation models: TabPFN-3~\cite{grinsztajnTabPFN3Technical2026} and ConTextTab~\cite{spinaciConTextTabSemanticsAwareTabular2025a}.
For both models, we use one classifier for NA and two separate regressors for RT and NT, resulting in three independent predictors. TabPFN-3 results are unavailable for BPI17 because evaluation could not be completed due to computational constraints at this dataset scale.

\mypar{Hyperparameters}
Hyperparameters are selected separately for each model family. For the LSTM baseline, we vary the number of layers, embedding size, hidden size, learning rate, and batch size, and train all configurations for 25 epochs. For the Transformer baseline trained from scratch, we vary the same parameters using a separate model-specific grid and additionally select the number of epochs from ${10,25}$. For the decoder-only LLM backbones, the pretrained architecture remains fixed and only the LoRA configuration is tuned, using rank $r\in{16,32,64,128,256}$ and scaling parameter $\alpha=2r$. The LoRA-adapted LLMs are trained for 10 epochs. We use LoRA fine-tuning for the main LLM comparison because it outperformed zero-shot and few-shot configurations in preliminary experiments. The tabular foundation models use fixed model-specific settings and do not undergo gradient-based training. All experiments are repeated with five pseudo-random seeds.

\mypar{Metrics}
NA is evaluated using accuracy and optimized using cross-entropy loss.
The continuous RT and NT tasks are evaluated using mean squared error (MSE) and optimized using MSE loss.
Higher values indicate better performance for accuracy, whereas lower values indicate better performance for MSE.
Computational efficiency is reported as runtime, defined as the wall-clock time elapsed between the start and end of each experimental run, reported in minutes. For the slice-level temporal analysis, we additionally report MAE to reduce sensitivity to large errors.

\subsection{Results}
\label{sec:results}

\mypar{Overall results}
Table~\ref{tab:baseline_vs_lora} reports the benchmark results for all three modeling paradigms across the NA, RT, and NT tasks on five event logs, including wall-clock runtime. Overall, no paradigm consistently dominates across all logs and tasks. Nevertheless, several notable patterns emerge.
First, the sequence models, LSTM and Transformer, achieve the highest NA accuracy on all five logs. In several cases, their advantage is substantial. For instance, on BPI12, the LSTM reaches an accuracy of 78\%, whereas TabPFN-3 and Gemma achieve only 64\%. The weaker performance of tabular models is particularly pronounced on the large BPI17 log: TabPFN-3 cannot process the full log due to its context-size limitation, while ConTextTab trails the Transformer by 21 percentage points.
For the temporal prediction tasks, the picture is more mixed. On RT prediction, tabular models achieve the best performance most frequently. On NT prediction, sequence and tabular models each obtain the best result on two out of five logs. In contrast, the LLM-based models generally lag behind on the temporal tasks.

\begin{table}[t!]
\centering
\caption{Results per dataset and model. Best values are highlighted in bold.}
\label{tab:baseline_vs_lora}
\scriptsize
\setlength{\tabcolsep}{1pt}
\renewcommand{\arraystretch}{1.05}
\begin{tabularx}{\textwidth}{>{\centering\arraybackslash}p{0.55cm} >{\raggedright\arraybackslash}p{2.1cm} *{4}{>{\centering\arraybackslash}X}}
\toprule
\multicolumn{1}{l}{\textbf{Dataset}} &
\multicolumn{1}{l}{\textbf{Model}} &
\multicolumn{1}{c}{\textbf{NA Acc.}} &
\multicolumn{1}{c}{\textbf{RT MSE}} &
\multicolumn{1}{c}{\textbf{NT MSE}} &
\multicolumn{1}{c}{\makecell[c]{\textbf{Runtime}\\\textbf{(min)}}} \\
\midrule
\multirow{7}{*}{\rotatebox[origin=c]{90}{BPI20PTC}}
& Majority       & 0.12 & 1.59 & 1.59 & \textbf{0.02} \\
& LSTM           & \textbf{0.78 $\pm$ 0.02} & 1.07 $\pm$ 0.03 & 1.19 $\pm$ 0.05 & 0.19 \\
& Transformer    & 0.78 $\pm$ 0.01 & 1.12 $\pm$ 0.09 & 1.24 $\pm$ 0.05 & 0.10 \\
& TabPFN-3       & 0.65 $\pm$ 0.001 & \textbf{1.02 $\pm$ 0.00} & 1.16 $\pm$ 0.00 & 0.60 \\
& ConTextTab     & 0.76 $\pm$ 0.00 & 1.05 $\pm$ 0.00 & \textbf{1.09 $\pm$ 0.00} & 0.16 \\
& Llama-3.2-1B   & 0.68 $\pm$ 0.18 & 1.05 $\pm$ 0.16 & 1.16 $\pm$ 0.18 & 4.82 \\
& Gemma-2-2b     & 0.62 $\pm$ 0.02 & 1.23 $\pm$ 0.50 & 1.23 $\pm$ 0.32 & 7.93 \\
\midrule

\multirow{7}{*}{\rotatebox[origin=c]{90}{BPI20RfP}}
& Majority & 0.17 & 1.15 & 1.02 & 0.35  \\
& LSTM & 0.88 $\pm$ 0.02 & 0.68 $\pm$ 0.02 & 0.75 $\pm$ 0.01 & 0.57  \\
& Transformer & \textbf{0.89 $\pm$ 0.00} & 0.66 $\pm$ 0.01 & 0.74 $\pm$ 0.01 & 0.28  \\
& TabPFN-3 & 0.73 $\pm$ 0.00 & 0.66 $\pm$ 0.02 & \textbf{0.70 $\pm$ 0.05} & 1.43  \\
& ConTextTab & 0.74 $\pm$ 0.01 & \textbf{0.64 $\pm$ 0.02} & 0.71 $\pm$ 0.06 & \textbf{0.17}  \\
& Llama-3.2-1B & 0.88 $\pm$ 0.02 & 0.69 $\pm$ 0.02 & 0.78 $\pm$ 0.03 & 10.15  \\
& Gemma-2-2b & 0.88 $\pm$ 0.01 & 0.69 $\pm$ 0.02 & 0.73 $\pm$ 0.01 & 23.31  \\
\midrule

\multirow{7}{*}{\rotatebox[origin=c]{90}{BPI20TPD}}
& Majority          & 0.08 & 1.46 & 1.25 & 0.40 \\
& LSTM              & \textbf{0.73 $\pm$ 0.00} & 1.01 $\pm$ 0.04 & 1.05 $\pm$ 0.03 & 0.70 \\
& Transformer       & 0.72 $\pm$ 0.00 & \textbf{0.81 $\pm$ 0.04} & \textbf{0.87 $\pm$ 0.03} & 0.48 \\
& TabPFN-3          & 0.61 $\pm$ 0.01 & 1.02 $\pm$ 0.03 & 1.03 $\pm$ 0.05 & 4.81 \\
& ConTextTab        & 0.68 $\pm$ 0.02 & 0.96 $\pm$ 0.06 & 1.03 $\pm$ 0.05 & \textbf{0.21} \\
& Llama-3.2-1B      & 0.70 $\pm$ 0.04 & 0.95 $\pm$ 0.07 & 0.95 $\pm$ 0.05 & 12.41 \\
& Gemma-2-2b        & 0.64 $\pm$ 0.01 & 1.05 $\pm$ 0.08 & 1.02 $\pm$ 0.07 & 24.14 \\
\midrule

\multirow{7}{*}{\rotatebox[origin=c]{90}{BPI12}}
& Majority      & 0.23          & \textbf{1.39}         & 1.54          & 0.59 \\
& LSTM          & \textbf{0.78 $\pm$ 0.01}  & 1.77 $\pm$ 0.09    & \textbf{1.34 $\pm$ 0.03}  & 1.17 \\
& Transformer   & 0.76 $\pm$ 0.01    & 2.74 $\pm$ 0.11    & 1.43 $\pm$ 0.02    & 1.13 \\
& TabPFN-3      & 0.64 $\pm$ 0.01   & 2.28 $\pm$ 0.06   & 1.60 $\pm$ 0.28   & 5.48 \\
& ConTextTab    & 0.63 $\pm$ 0.01   & 2.15 $\pm$ 0.10    & 1.64 $\pm$ 0.32    & \textbf{0.29} \\
& Llama-3.2-1B  & 0.48 $\pm$ 0.11    & 2.68 $\pm$ 0.13    & 1.66 $\pm$ 0.04    & 29.10 \\
& Gemma-2-2b    & 0.64 $\pm$ 0.05    & 2.34 $\pm$ 0.25    & 1.47 $\pm$ 0.09    & 56.53 \\
\midrule

\multirow{7}{*}{\rotatebox[origin=c]{90}{BPI17}}
& Majority      & 0.15       & 1.10       & 1.21      & 1.79 \\
& LSTM          & 0.85 $\pm$ 0.00 & 0.67 $\pm$ 0.03 & 0.76 $\pm$ 0.02 & 2.50 \\
& Transformer   & \textbf{0.86 $\pm$ 0.01} & 1.16 $\pm$ 0.08 & 0.88 $\pm$ 0.03 & 1.17 \\
& TabPFN-3      & -- & -- & -- & -- \\
& ConTextTab    & 0.65 $\pm$ 0.02 & 1.30 $\pm$ 0.25 & 1.12 $\pm$ 0.16 & \textbf{0.71} \\
& Llama-3.2-1B  & 0.80 $\pm$ 0.04 & \textbf{0.66 $\pm$ 0.03} & \textbf{0.73 $\pm$ 0.02} & 112.76 \\
& Gemma-2-2b    & 0.81 $\pm$ 0.04 & 1.30 $\pm$ 0.13 & 0.94 $\pm$ 0.06 & 232.53 \\
\bottomrule
\end{tabularx}

\emph{Note:} An en dash indicates an unavailable result. TabPFN-3 could not be evaluated on BPI17 due to computational constraints at this dataset scale.
\end{table}

\mypar{Runtime}
Sequence models and ConTextTab exhibit substantially lower recorded runtimes than the LoRA-fine-tuned LLMs, whose runtime increases markedly with log size. TabPFN-3 is consistently slower than ConTextTab and, on several logs, also slower than the sequence baselines. Thus, the absence of gradient-based training does not automatically imply lower runtime. The reported values exclude preprocessing and setup overhead and should be interpreted as indicative rather than end-to-end efficiency estimates.

\begin{figure}[t]
    \centering
    \includegraphics[width=0.82\textwidth]{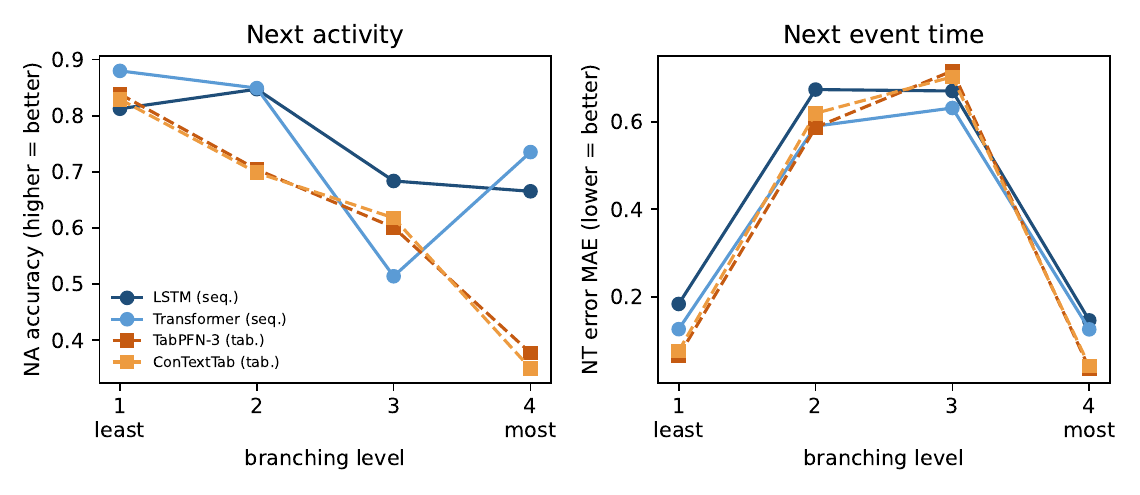}
    \caption{Illustrative branching analysis for BPI12 based on one representative seed. Sequence and tabular models separate more strongly for next activity prediction as branching increases, whereas no comparable separation is visible for next event time prediction.}
    \label{fig:branching_dissociation}
\end{figure}

\mypar{Discussion}
Figure~\ref{fig:branching_dissociation} illustrates the slice-level analysis for BPI12. For NA prediction, the sequence--tabular gap generally increases with branching, while tabular models often remain competitive on prefixes with limited branching. Here, branching refers to the ambiguity of the next step at the current activity, quantified by the number and distribution of observed successor activities. This pattern is consistent with the representational distinction discussed in Section~\ref{sec:models}: tabular FMs encode prefixes as fixed-length state vectors rather than ordered trajectories, while sequence models retain the event order. The pattern is less pronounced for the temporal targets. For RT, the sequence advantage largely remains flat or decreases as branching increases. For NT, the results are more heterogeneous, with a weaker and less consistent branching-related trend than for NA. The branching-related disadvantage of tabular models is therefore most clearly pronounced for NA.

Although LLM-based approaches preserve the sequential prefix representation, they remain affected by branching-related difficulty. Unlike the broadly lower NA performance of tabular models, the LLM deficit is concentrated on selected logs. This suggests dataset-dependent adaptation quality rather than a uniform representational limitation. BPI12 provides a concrete example: Llama-3.2-1B and Gemma-2-2b frequently predict the end-of-sequence token \texttt{} before the process has actually ended. These premature \texttt{} predictions account for 23\% and 25\% of their NA errors, respectively, compared with 0\% for the LSTM and Transformer baselines.

\section{Conclusion}
\label{sec:conclusion}

Across five event logs and three predictive tasks, we compared sequence models trained from scratch, fine-tuned LLMs, and tabular foundation models.

The sequence models achieved the strongest NA accuracy across all evaluated logs, demonstrating the effectiveness of learning process-specific sequential patterns directly from event data. While tabular foundation models clearly fall behind on NA prediction, they show competitive performance on the temporal tasks. LLMs, which incurred substantially higher runtimes due to fine-tuning, showed stronger performance than the tabular models on NA, but usually lag behind on the temporal targets. A further analysis indicates that the sequence--tabular performance gap for NA is driven by increasing branching complexity.

To obtain a more comprehensive understanding, future work should examine the effectiveness of the different paradigms across additional PPM tasks and a broader range of process characteristics. Moreover, the evaluated LLMs are comparatively small; incorporating larger models may provide further insights into the performance potential of this paradigm. Finally, this work does not investigate the impact of alternative feature representations, which may be particularly relevant for tabular models to better capture trace-level structure.

%
%
\bibliographystyle{splncs}
\bibliography{bib}

\end{document}